\documentclass[10pt,twocolumn,letterpaper]{article}

\usepackage{cvpr}              

\usepackage[dvipsnames]{xcolor}

\definecolor{cvprblue}{rgb}{0.21,0.49,0.74}
\usepackage[pagebackref,breaklinks,colorlinks,citecolor=cvprblue]{hyperref}

\def\paperID{*****} 
\def\confName{CVPR}
\def\confYear{2024}

\title{Cross-Domain Learning for Video Anomaly Detection with Limited Supervision}

\author{Yashika Jain\\
University of Delhi\\
{\tt\small yashikajain201@gmail.com}
\and
Ali Dabouei \thanks{Corresponding authors.}\\
Carnegie Mellon University\\
{\tt\small ali.dabouei@gmail.com}
\and
Min Xu \protect\footnotemark[1]\\
Carnegie Mellon University\\
{\tt\small mxu1@cs.cmu.edu}
}

\usepackage{graphicx}
\usepackage{booktabs}
\usepackage{times}
\usepackage{epsfig}
\usepackage{graphicx}
\usepackage{amsmath}
\usepackage{amssymb}
\usepackage{graphicx}
\usepackage{booktabs}
\usepackage{array}
\usepackage{dblfloatfix}
\usepackage{tabularx}
\usepackage{amsfonts}
\usepackage{siunitx}
\usepackage{multirow}
\usepackage{multicol}
\usepackage{cellspace}
\usepackage{pifont}
\usepackage{caption}
\usepackage{accents}
\usepackage{bm}
\usepackage{wrapfig}
\usepackage{float}
\usepackage{adjustbox}

\def\bZm{\mathbf{Z}_{\text{m}}}
\def\bZa{\mathbf{Z}_{a}}

\def\bzm{\mathbf{z}_{\text{m}}}
\def\bza{\mathbf{z}_{a}}
\def\bs{\mathbf{s}}

\begin{document}
\maketitle
\begin{abstract}
Video Anomaly Detection (VAD) automates the identification of unusual events, such as security threats in surveillance videos. In real-world applications, VAD models must effectively operate in cross-domain settings, identifying rare anomalies and scenarios not well-represented in the training data.
However, existing cross-domain VAD methods focus on unsupervised learning, resulting in performance that falls short of real-world expectations. Since acquiring weak supervision, \ie, video-level labels, for the source domain is cost-effective, we conjecture that combining it with external unlabeled data has notable potential to enhance cross-domain performance.
To this end, we introduce a novel weakly-supervised framework for Cross-Domain Learning (CDL) in VAD that incorporates external data during training by estimating its prediction bias and adaptively minimizing that using the predicted uncertainty. We demonstrate the effectiveness of the proposed CDL framework through comprehensive experiments conducted in various configurations on two large-scale VAD datasets: UCF-Crime and XD-Violence. Our method significantly surpasses the state-of-the-art works in cross-domain evaluations, achieving an average absolute improvement of 19.6\% on UCF-Crime and 12.87\% on XD-Violence.
\end{abstract}    
\section{Introduction}
\label{sec:intro}

Video anomaly detection (VAD) aims to locate anomalous events in the videos \cite{sultani2018real, tian2021weakly, zhong2019graph, feng2021mist, liu2018future, Hasan_2016_CVPR, cong2011sparse, Zhang_2023_CVPR, Aich_2023_WACV, Lv_2021_CVPR}. Unlike manual surveillance, which is costly and time-consuming, video anomaly detection eliminates the need for extensive human effort, saving resources and time. 
It holds significant potential for playing a vital role in video surveillance by identifying unusual behaviors and activities such as accidents, burglaries, explosions, and other events that signal security threats.

VAD has been extensively studied previously \cite{sultani2018real, tian2021weakly, zhong2019graph, feng2021mist, liu2018future, Hasan_2016_CVPR}. Owing to the high costs and time associated with obtaining frame-level labels, most approaches formulate the problem as either an unsupervised \cite{liu2018future, Hasan_2016_CVPR, cong2011sparse} or weakly-supervised learning setup \cite{sultani2018real, tian2021weakly, feng2021mist}.
In the unsupervised or one-class classification-based) learning setup, only normal videos are used to model the underlying distribution of normal spatiotemporal patterns, and any deviations from the modeled distribution are regarded as anomalies. Despite the convenience of the unsupervised setup, the lack of anomalous videos during training limits the model's ability to learn the specific characteristics of anomalies. This results in limited performance which does not meet real-world expectations. 
To address this issue, weakly-supervised setup has attracted significant attention. In this setup, merely video-level labels indicating the presence of anomalies within the videos are incorporated as weak supervision to train models capable of making frame-level predictions at inference.
Multiple Instance Learning (MIL) \cite{sultani2018real} is a prominent technique in this domain. By treating each video as a ``bag'' and each snippet as a ``segment'',  MIL-based algorithms operate under the premise of a worst-case scenario where the segment with the highest predicted probability of being abnormal is considered as the candidate to represent the whole video.

\begin{figure}[t]
    \centering
    \includegraphics[width=0.43\textwidth]{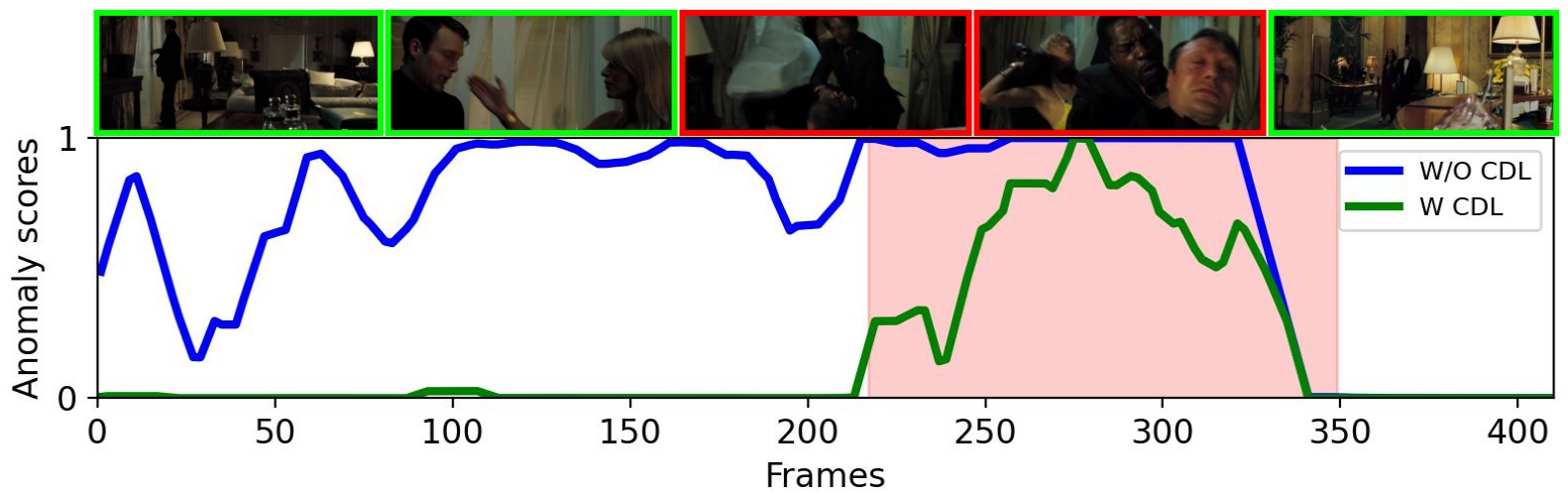}
    \caption{Anomaly score comparison on a video of XD-Violence dataset, with and without employing the proposed CDL framework. The model trained without CDL on UCF-Crime as the weakly labeled set consistently yields high anomaly scores. In contrast, the model trained with CDL, using UCF-Crime as the weakly labeled set and HACS as the unlabeled set, is better able to localize the anomalous frames.}   
    \label{fig:intro-figure}
\end{figure}

In real-world applications, it is inevitable to encounter environments and scenarios not fully represented in the model's training set. However, it is essential that the model makes correct predictions in such novel situations. For instance, when the training data lacks samples of rare events like ``riots'' or accidents in novel scenes, the model should be able to characterize such occurrences as anomalous when they occur. Previous works study these novel situations under the cross-domain problem definition \cite{9410375, Aich_2023_WACV, 10.1007/978-3-030-58558-7_8}. 

Existing cross-domain VAD methods \cite{9410375, Aich_2023_WACV, Lv_2021_CVPR, 10.1007/978-3-030-58558-7_8} rely on unsupervised techniques and consequently exhibit limited performance, as demonstrated later in our empirical evaluations in Tables \ref{table:cdl-ucf} and \ref{table:cdl-xdv}. A solution to this could be the adoption of weakly-supervised techniques for cross-domain VAD. While weakly-supervised approaches have proven promising in single-domain scenarios \cite{sultani2018real, tian2021weakly, feng2021mist}, their effectiveness in cross-domain scenarios has not been extensively explored. Our evaluations in Tables \ref{table:cdl-ucf} and \ref{table:cdl-xdv} suggest that directly employing existing weakly-supervised methods to address the cross-domain challenges results in a significant performance drop when tested in scenarios of even similar nature, such as surveillance videos. We argue that this performance gap is due to the following reasons. First, anomalous events, by their very nature, lack a specific pattern or predefined structure. Hence, the definition of anomaly is context-dependent and a naive adaptation of the previous method cannot capture the context-dependencies in multiple domains. Second, anomalous events are relatively infrequent, making VAD a class imbalance problem. This issue becomes more severe when dealing with multiple domains. Third, because of the limited amount of weakly labeled training data, the model's learning capacity to detect novel (open-set) anomalies is also constrained.
Due to these challenges, weakly-supervised methods cannot be readily applied to cross-domain or cross-dataset scenarios. 

To overcome these challenges and develop a generalized VAD model, substantial amounts of weakly-labeled data are required. However, acquiring even video-level labels for a large number of videos is inefficient and labor-intensive. On the other hand, vast streams of unlabeled videos are generally available. Utilizing the limited weakly-labeled data alongside this abundant unlabeled data provides a notable opportunity to address the aforementioned challenges in cross-domain VAD. Prudent utilization of the unlabeled data can provide valuable insights into the underlying data distribution, leading to improved decision-making and identification of anomalous events.

To this end, we propose a weakly-supervised \textit{Cross-Domain Learning (CDL) framework} for VAD that integrates external, unlabeled data, from the wild with limited weakly-labeled data to provide competitive generalization across the domains. This is achieved by adaptively minimizing the prediction bias over the external data using the estimated prediction variance, which serves as an uncertainty regularization score. In the proposed framework, we first train fine-grained pseudo-label generation models on the weakly-labeled data to obtain sets of segment-level predictions for the external dataset. Second, we compute the variance of the predictions across multiple predictors as a proxy to represent uncertainty associated with the segments in the external data. Third, during the optimization process, involving training on both labeled and external data, we adaptively reweigh the bias on each external data using the uncertainty regularization scores. This dynamic reweighing ensures that segments from the external dataset closer to the source dataset are emphasized during the training, while those with higher uncertainty are down-weighted. Finally, we iteratively regenerate pseudo-labels using the models trained on labeled and pseudo-labeled data, re-estimate the uncertainties, and re-train the model on the union of labeled and external datasets. This iterative process helps refine the pseudo-labels as the training progresses. With this training process, the model learns to generalize to both source and external data, given only supervision on the source data. Figure \ref{fig:intro-figure} illustrates the effectiveness of the CDL framework.

To summarize, we make the following contributions:

\begin{itemize}
    \item We present a practical CDL framework for weakly-supervised VAD, in which unlabeled external videos are employed to enhance the cross-domain generalization of the model.
    \item We design a novel uncertainty quantification method that enables the adaptive uncertainty-driven integration of external videos into the training set.
    \item Through extensive experiments and ablation studies on benchmark datasets, we validate the proposed approach, demonstrating state-of-the-art performance in cross-domain settings while retaining a competitive performance on the in-domain data.
\end{itemize}

\begin{table}[t]
\small
    \centering
    \begin{tabular}{lcc}
        \toprule
        Method(s) &  Sup. on $\mathcal{D}$ & Target \\
        \midrule
        Acsintoae \textit{et al.} \cite{Acsintoae_CVPR_2022} & unsupervised & $\mathcal{D}$ \\
        rGAN \cite{10.1007/978-3-030-58558-7_8}, MPN \cite{Lv_2021_CVPR} & unsupervised & $\mathcal{D}'$ \\
        zxVAD \cite{Aich_2023_WACV} & unsupervised & $\mathcal{D}\cup\mathcal{D}'$ \\ 
        Ours & weakly-supervised &  $\mathcal{D}\cup\mathcal{D}'$ \\
        \bottomrule
    \end{tabular}
    \caption{Brief overview of the taxonomy of current works for VAD using a source domain dataset ($\mathcal{D}$) and a secondary domain dataset ($\mathcal{D}'$). All these methods do not utilize any labels for training on ($\mathcal{D}'$) and assume distinct distributions for $\mathcal{D}$ and $\mathcal{D}'$.}
    \label{table:current-works}
\end{table}

\section{Related Works}
\label{section:related-works}
\textbf{Video Anomaly Detection (VAD).}
\textbf{Video Anomaly Detection (VAD).}
VAD is a well-established problem, with most works formulating it either as unsupervised learning \cite{liu2018future, 6751449, inproceedings, yu2020cloze, Hasan_2016_CVPR} or weakly-supervised learning \cite{sultani2018real, 8803657, zhu2019motion, tian2021weakly, sapkota2022bayesian} problem. In unsupervised setups, the training data consists solely of normal videos, with the majority of works encoding normal patterns through techniques like frame reconstruction \cite{Hasan_2016_CVPR, xu2015learning}, future frame prediction \cite{liu2018future}, dictionary learning \cite{6751449, inproceedings}, and one-class classification \cite{8019325, ionescu2019object}. Any deviation from the encoded patterns is considered anomalous. Since the model categorizes anything beyond its learned representations as anomalous, it can label novel video actions and scenarios encountered during training but in altered environments as anomalous.
Weakly-supervised VAD methods help mitigate these issues by incorporating video-level labels as weak supervision for the model, with the majority of methods utilizing the Multiple Instance Ranking Loss \cite{sultani2018real, feng2021mist, Wan2020WeaklySV, zhong2019graph}.
 \begin{figure*}[tb]
    \centering
    \includegraphics[width=1\textwidth]{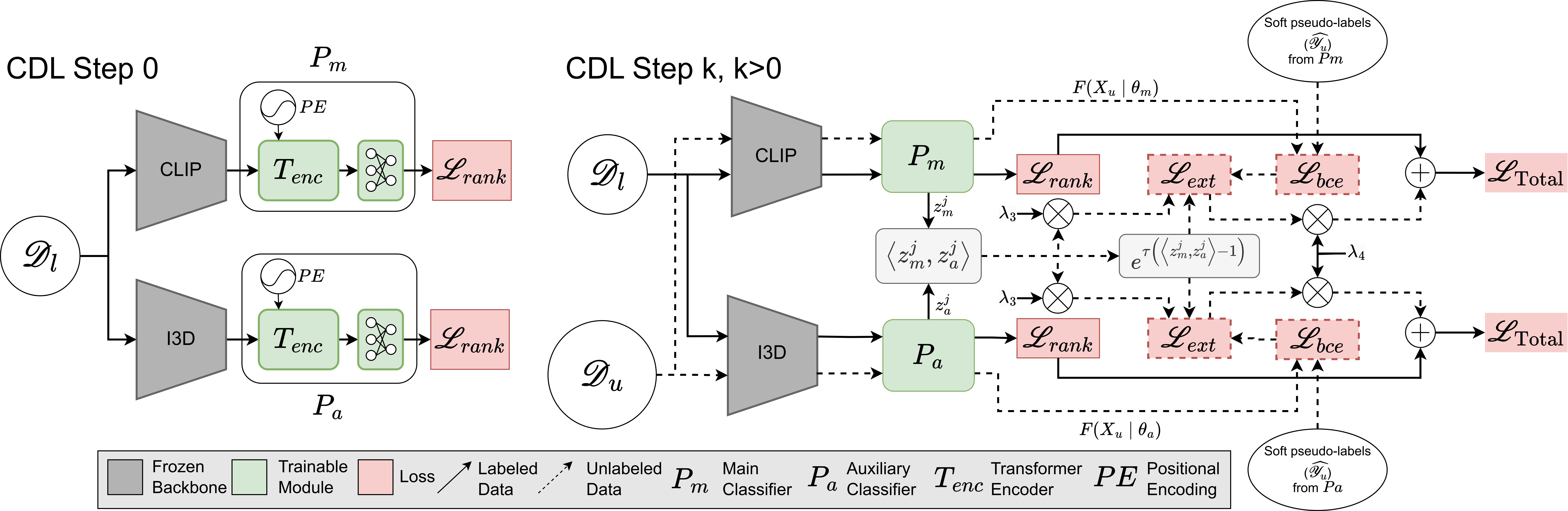}
    \caption{
Overview of the proposed CDL Framework. \textbf{CDL Step 0:} The Ranking Loss, $\mathcal{L}_{\text{rank}}$ (Supp Mat. \textsection 6), is employed to train two pseudo-label generation models, $P_m$ and $P_a$, \textsection \ref{sec:extraction}, on weakly-labeled data, $\mathcal{D}_l$. \textbf{CDL Step} \bm{$k, k>0$}: $P_m$ and $P_a$ are trained iteratively on $\mathcal{D}_l \cup \mathcal{D}_u$, incorporating pseudo-labels for $\mathcal{D}_u$ generated at the end of the previous CDL step. To deal with noise in pseudo-labels, uncertainty regularization scores are estimated using the divergence between the predictions of the two models, \textsection \ref{sec:uncertainty}. When optimizing on $\mathcal{D}_u$, the prediction bias, $\mathcal{L}_{\text{bce}}$ (\textsection \ref{sec:prediction-bias}), for external data is reweighed using the computed uncertainty regularization scores, \textsection \ref{sec:optimization}.
}
\label{fig:method}
\end{figure*}  
Given that a VAD model is expected to encounter previously unseen scenarios during deployment, it is of paramount importance for the model to have a high generalization across domains. Previous works refer this as cross-domain \cite{Aich_2023_WACV} or cross-dataset generalization \cite{Cho_2023_CVPR}.
We provide an overview of the existing works employing external data in VAD in Table \ref{table:current-works}. Previous works on cross-domain generalization focus on unsupervised methods based on few-shot target-domain scene adaptation. \cite{10.1007/978-3-030-58558-7_8, Lv_2021_CVPR} employ data from the target domain via meta-learning to adapt to that specific domain.
Aich \textit{et al.} \cite{Aich_2023_WACV} proposed a zero-shot target domain adaptation method that incorporates external data to generate pseudo-abnormal frames. 
Despite the intriguing setup, these unsupervised cross-domain generalization methods lack explicit knowledge about what constitutes an anomaly, hindering the model's ability to learn the specific characteristics of anomalies.
To this end, we propose the use of weakly-supervised learning for cross-domain generalization. We integrate external datasets from diverse domains to enable the cross-domain generalization of a model trained in a weakly-supervised fashion.\\
\textbf{Pseudo-Labeling and Self-training.}
Pseudo-labeling \cite{pseudoLabel2019, rizve2021in} is a common technique where the model trained on labeled data assigns labels to unlabeled data. 
Subsequently, the model is trained on both the initially labeled data and the pseudo-labeled data. This self-training strategy \cite{10.3115/981658.981684, mcclosky-etal-2006-effective} operates iteratively, allowing the model to progressively enhance its generalization. In VAD, several works leverage pseudo-labeling and self-training for generating fine-grained pseudo-labels \cite{feng2021mist, Li_Liu_Jiao_2022, Zhang_2023_CVPR}. However, in contrast to the previous methods, instead of generating pseudo-labels for the weakly labeled data, we leverage pseudo-labels for incorporating the external data. \\
\textbf{Uncertainty Estimation.} To address pseudo-label noise, prior research in different contexts has explored uncertainty estimation using various approaches, such as data augmentation \cite{10.5555/3495724.3495775, Berthelot2020ReMixMatch:}, inference augmentation \cite{pmlr-v48-gal16}, and model augmentation \cite{zheng2021rectifying}.  While data augmentation is effective for images, it can disrupt temporal relationships in video frames and is not efficient for training on high-cardinality data like videos. On the other hand, inference augmentation methods, such as MC Dropout \cite{pmlr-v48-gal16, Zhang_2023_CVPR}, introduce perturbations during model inference to obtain slightly different predictions, but that is inefficient for training with fixed backbones. In contrast, model augmentation uses different models. Since different models may have varying biases and receptive fields, this would result in diverse predictions. This prediction discrepancy can help quantify uncertainty, making model augmentation well-aligned with our problem. To avoid any manual thresholding for learning from pseudo-labels during training, following \cite{pmlr-v180-huang22a, zheng2021rectifying} we use adaptive reweighing of loss with uncertainty values. 
In \cite{zheng2021rectifying}, Zheng \textit{et al.} quantify uncertainty by estimating discrepancies between predictions made by two classifiers using Kullback–Leibler (KL) divergence. However, given that VAD is a binary classification task, the divergence based on only two outcomes for the posterior probability is not optimally informative. Hence, we propose a method to quantify uncertainty in the high-dimensional feature space instead of the probability space.

\section{Method}
\subsection{Problem Definition}\label{sec:problem-def}
In this work, we address a real-world VAD problem, where a weakly-labeled dataset $\mathcal{D}_{l}=\{(X_{l}^{i}, Y_{l}^{i})\}_{i=1}^{n_l}$ and an external unlabeled dataset $\mathcal{D}_{u}=\{X_{u}^{i}\}_{i=1}^{n_u}$ are available for training. Here, $n_l$ and $n_u$ indicate the number of videos in the two datasets, respectively, with $n_u \gg n_l$ due to the convenience of gathering unlabeled video data. The video-level labels of $X_{l}$ are denoted by $Y_{l} \in \{0, 1\}$. We do not make any assumption about distributions of $\mathcal{D}_l$ and $\mathcal{D}_u$, and therefore, they can be drawn from different distributions. 
We aim to find the model $F(\cdot|\theta)$, parameterized by $\theta$, that provides accurate predictions on weakly-labeled data while adaptively minimizing the prediction bias on the external data using the uncertainty regularization scores. We illustrate the proposed framework in Figure \ref{fig:method}.
\subsection{Feature Extraction and Temporal Processing}\label{sec:extraction}
The proposed uncertainty quantification method (Section \ref{sec:uncertainty}) compares two diverse representations of each sample to estimate the uncertainty associated with the segment-level predictions on external data. To this aim, we employ two different backbones for feature extraction from videos, which are widely used for anomaly detection tasks.
The first one is the conventional I3D backbone \cite{carreira2017quo}, which extracts segment-level features using 3D convolution, and the other is the CLIP backbone \cite{pmlr-v139-radford21a}, which extracts frame-level features using the frozen CLIP Model's ViT encoder. 
The contrasting inductive biases of the 3D convolution-based I3D and the transformer-based CLIP help to effectively capture the prediction variance. It is to be noted that only the CLIP backbone is used during inference. We develop two prediction heads, namely the main model, $P_m$, built on top of the CLIP backbone, and the auxiliary model, $P_a$, built on top of the I3D backbone.

Video frames are highly correlated in the temporal dimension. To reduce the redundancy in frame-level features extracted by the CLIP backbone, we pool the representations by bilinearly interpolating them to a fixed, empirically determined length, $n_s$. Each of the $n_s$ interpolated features represents one segment. To ensure consistency, we also fix the length of representations extracted by the I3D backbone. Evaluation in Section \ref{sec:ablation-studies} analyzes the role of $n_s$ on the model's performance. 
To capture long-range temporal information over the sequence, we employ a lightweight temporal network, \ie, transformer encoder, to implement ${P_m}$ and ${P_a}$.
\subsection{Bias Estimation for External Data} \label{sec:prediction-bias}
Similar to \cite{zheng2021rectifying}, we formulate the prediction bias on external data as:
\begin{equation}
{Bias}(\mathcal{D}_u) = \mathbb{E}_{X_u}[F(X_{u}|\theta) - \mathcal{Y}_{u}],
\end{equation}  
where $F(X_{u}|\theta)$ represents a set of predicted probability distributions, each one corresponding to a distinct segment of $X_{u}$, and $\mathcal{Y}_{u}$ denotes the set of unknown segment-level labels of $X_{u}$.
${Bias}({\mathcal{D}_u})$ can be re-written as:
\begin{equation}  \label{eq:2}
{Bias}(\mathcal{D}_u) = \mathbb{E}_{X_u}[F(X_{u}|\theta) - \mathcal{\hat{Y}}_{u}] + \mathbb{E}_{X_u}[ \mathcal{\hat{Y}}_{u} - \mathcal{Y}_{u}],
\end{equation} 
where $\mathcal{\hat{Y}}_{u}$ denotes the set of segment-level pseudo-labels for $X_{u}$. $\mathcal{\hat{Y}}_{u}$ can be generated by performing inference on the model trained on $\mathcal{D}_{l}$. The first term in Equation \ref{eq:2} denotes the difference between the predicted posterior probability and the pseudo-labels, while the second term denotes the error between the pseudo-labels and the ground-truth labels. 
While minimizing the prediction bias, due to the lack of ground truth supervision, we employ a self-training mechanism, considering $\mathcal{\hat{Y}}_{u}$ as the soft labels, thereby treating the second term as a constant and minimizing the first term.
Specifically, we use the binary cross-entropy (BCE) loss, $\mathcal{L}_{\text{bce}}$, given by:
\begin{equation}
\mathcal{L}_{\text{bce}} =  -\mathcal{\hat{Y}}_{u}\log(F(X_{u}|\theta)) \
 - (1 - \mathcal{\hat{Y}}_{u}) \log(1 - F(X_{u}|\theta)),
\end{equation}
to estimate the prediction bias associated with each video segment, for both ${P_m}$ and ${P_a}$. 

\subsection{Uncertainty Estimation}\label{sec:uncertainty}
Since ${\mathcal{D}_u}$ and $\mathcal{D}_l$ do not necessarily share the same distribution, the generated pseudo-labels are noisy. This noise can adversely affect the subsequent training process as it causes bias to further magnify and propagate within the model. This issue, known as Confirmation Bias \cite{pseudoLabel2019}, is often mitigated by quantifying the uncertainty associated with pseudo-labels and then incorporating this uncertainty into the training process to compensate for the noise.
As discussed in Section \ref{section:related-works}, we opt to address the confirmation bias by computing uncertainty using model augmentation.
To quantify uncertainty through model augmentation, following \cite{zheng2021rectifying}, we estimate prediction variance, which is formulated as:
\begin{equation} \label{var}
{Var}(\mathcal{D}_u) = \mathbb{E}_{X_u}[(F(X_{u}|\theta) - \mathcal{Y}_{u})^2].
\end{equation}  
Due to the lack of ground-truth labels, Equation \ref{var} can be approximated as:
\begin{equation} \label{var-approx}
{Var}(\mathcal{D}_u) \approx \mathbb{E}_{X_u}[(F(X_{u}|\theta) - \mathcal{\hat{Y}}_{u})^2].
\end{equation} 
When optimizing the prediction bias in Equation \ref{eq:2}, the variance in Equation \ref{var-approx} will also be minimized, potentially resulting in inaccurate quantification of the true prediction variance. To address this, we adopt an alternative approximation, expressed as:
\begin{equation} \label{var-approx-2}
{Var}(\mathcal{D}_u) \approx \mathbb{E}_{X_u}[\big(P_{m}(X_{u}|\theta_{P_{m}}) - P_{a}(X_{u}|\theta_{P_{a}})\big)^2].  
\end{equation} 
Since VAD is a binary classification task, the probability distributions corresponding to each segment have limited support. Consequently, estimating prediction variance using only the predicted anomaly scores, as in Equation \ref{var-approx-2}, may not be robust.
Hence, instead of measuring the divergence between the predicted posterior probabilities for the two classes, we propose quantifying pseudo-label uncertainty in the high-dimensional space. To this end, we compute the cosine similarity between the segments in each set of the representations, $Z_m$ and $Z_a$, obtained from the penultimate layer of ${P_m}$ and ${P_a}$, respectively. Here, $Z_m= \{\bzm^1, \bzm^2, \ldots, \bzm^{n_s}\}$ and $Z_a= \{\bza^1, \bza^2, \ldots, \bza^{n_s}\}$.

To obtain a set of stabilized, segment-level uncertainty regularization scores within a bounded range from the computed cosine similarity, we introduce the following function. Let $S= \{\bs^1, \bs^2, \ldots, \bs^{n_s}\}$ be the set of surrogate variances that we use as proxies for the uncertainty of segments. The surrogate variance is computed as:
\begin{equation} \label{uncertainty}
\bs^j = e^{\tau(\langle\bzm^j, \bza^j\rangle - 1)},
\end{equation}
where $\bs^j$ indicates the uncertainty regularization score for the $j^{th}$ segment, ${\langle\bzm^j, \bza^j\rangle}$ indicates the cosine similarity, and $\tau$ denotes the temperature parameter.

Higher uncertainty regularization scores indicate the similar encoding of data between the models, implying less uncertainty in the predicted labels, while, lower scores imply high uncertainty in the predicted labels. Empirical evidence in Section \ref{sec:correlation} demonstrates a significant negative correlation between uncertainty regularization scores and Binary Cross-Entropy (BCE) loss between the predicted labels and ground truths. This affirms that the proposed uncertainty regularization score effectively serves as a proxy for the quality of pseudo-labels.

\subsection{Training Process} \label{sec:optimization}
\textbf{CDL Step 0}. We initially train ${P_m}$ and ${P_a}$ separately on the labeled set, optimizing both of them using the Ranking Loss, $\mathcal{L}_{\text{rank}}$, discussed in Supp. Mat. Sec. 6. We then perform inference on the trained models to generate the sets of soft segment-level pseudo-labels for training on ${\mathcal{D}_u}$. \\
\textbf{CDL Step $\textgreater$ 0}. Following the generation of the sets of pseudo-labels for ${\mathcal{D}_u}$, we enter an iterative pseudo-label refinement phase, where we train ${P_m}$ and ${P_a}$ on $\mathcal{D}_l \cup \mathcal{D}_u$ for multiple CDL steps. Each CDL step comprises a fixed number of epochs. In each epoch, we regenerate the sets of segment-level uncertainty regularization scores. To enable the uncertainty-driven learning from external data, similar to \cite{zheng2021rectifying}, we use the estimated uncertainty regularization scores, $S$, as automatic thresholds as this dynamically adjusts learning from noisy labels by scaling the prediction bias associated with external data based on $S$. This helps filter out unreliable predictions while prioritizing highly confident predictions. 
To encourage lower prediction variance, which would in turn lead to increased pseudo-label quality, we explicitly add the prediction variance to the optimization objective corresponding to the external data, $\mathcal{L}_{\text{ext}}$, as:
\begin{equation} \label{eq:8}
\mathcal{L}_{\text{ext}} = \mathbb{E}_{X_u}[\frac{1}{Var(\mathcal{D}_u)} \cdot {Bias}(\mathcal{D}_u) + {Var}(\mathcal{D}_u) ].
\end{equation} 
Equation \ref{eq:8} is rewritten with the approximated terms as:
\begin{equation} \label{equation:ext-loss-expectation}
\mathcal{L}_{\text{ext}} = \mathbb{E}_{X_u}[S \cdot \mathcal{L}_{\text{bce}} - \lambda_{3} \cdot {\langle Z_m, Z_a\rangle}].
\end{equation}  
Alternatively, Equation \ref{equation:ext-loss-expectation} can be rewritten as:
\begin{equation}
\mathcal{L}_{\text{ext}} = \frac{1}{n_s \cdot n_u} \sum_{i=1}^{n_u}\sum_{j=1}^{n_s} \left(S^{i, j} \cdot \mathcal{L}_{\text{bce}}^{i, j} - \lambda_{3} \cdot {\langle\bZm^{i, j}, \bZa^{i, j}\rangle}\right),
\end{equation}
where $\lambda_{3}$ is a hyper-parameter to balance the losses.
Similar to CDL step 0, to optimize the training on $\mathcal{D}_l$, we use $\mathcal{L}_{\text{rank}}$.
The total optimization objective for training on $\mathcal{D}_l \cup \mathcal{D}_u$ can be expressed as:
\begin{align} \label{eq:total-objective}
\mathcal{L}_{\text{Total}} =\mathcal{L}_{\text{rank}} + \lambda_{4} \cdot \mathcal{L}_{\text{ext}},
\end{align}
where $\lambda_{4}$ is a trade-off parameter for $\mathcal{L}_{\text{ext}}$.
We employ the optimization objective defined in Equation \ref{eq:total-objective} during training on $\mathcal{D}_l \cup \mathcal{D}_u$ for each epoch within every CDL step.
After each CDL step is completed, we re-generate the set of soft segment-level pseudo-labels using the models trained on $\mathcal{D}_l \cup \mathcal{D}_u$.
This iterative refinement process repeats $k$ times, where $k$ is a hyper-parameter determining the number of CDL steps. With each CDL step, the models' performance gets further refined as the pseudo-labels get iteratively improved.
\subsection{Inference - Extending Segment-level Scores to Frame-level Scores} \label{sec:inference}
During inference, we compute segment-level anomaly scores for the videos using ${P_m}$. Since we encounter long-untrimmed videos with varying numbers of frames, for extending the segment-level anomaly score to the frame level, for each video, we divide the total number of frames ${n_f}$ by the number of segments ${n_s}$ to obtain the number of frames per segment, ${n_{fs}}$. We assign the anomaly score of each segment to its consecutive frames. The first segment corresponds to the first ${n_{fs}}$ frames, and so forth until the ${(n_s - 1)}^{th}$ segment. For the last segment, its anomaly score is assigned to any remaining frames, potentially exceeding ${n_{fs}}$, if there is a remainder.
\section{Experiments} \label{sec:Experiments}
We evaluate the proposed method on the major video anomaly datasets, UCF-Crime (UCF) \cite{sultani2018real} and XD-Violence (XDV) \cite{Wu2020not}. Additionally, we use 11,000 videos from the HACS \cite{zhao2019hacs} dataset as a source of external data.
We provide detailed information about the datasets in Supp. Mat. \textsection 7. In \textsection \ref{subsec:imp-details}, we discuss the implementation details. In \textsection \ref{subsec:noise}, we discuss the inherent noise in the test annotations of benchmark datasets.
We proceed to compare the proposed framework with prior works in cross-domain scenarios (\textsection \ref{subsubsec:cross-domain}) and open-set scenarios (\textsection \ref{subsubsec:open-set}). 
Subsequently, in \textsection \ref{sec:correlation}, we demonstrate a strong correlation between the quality of pseudo labels and the computed uncertainty scores.
We then explore the evolution of these uncertainty scores through the training process in \textsection \ref{subsec:progression}.
Finally, in \textsection \ref{sec:ablation-studies}, we conduct ablation studies and hyper-parameter analysis to analyze the impact of individual components of the proposed framework. 

\subsection{Implementation Details} \label{subsec:imp-details}
We implement the proposed method using PyTorch. We extract CLIP and I3D features at a fixed frame rate of 30 FPS. CLIP features are extracted from the frozen CLIP model's image encoder (ViT-B/32). For the hyper-parameters, in the open-set scenarios, we empirically set the value of $n_s$ to 64, $\tau$ to 1.25, $\lambda_{1}$ and $\lambda_{2}$ to $5e-4$, $\lambda_{3}$ to $1e-3$, and $\lambda_{4}$ to 700. Ablation studies for selecting $n_s$ and $\lambda_{3}$ are included in Section \ref{sec:ablation-studies}. We use the Adam optimizer with a weight decay of $1e-3$, and we set a learning rate of $3e-5$ for the transformer encoder and $5e-4$ for the fully connected layers. We use a batch size of 64. In both ${P_m}$ and ${P_a}$, we explicitly encode positional information in the segments using sinusoidal positional encodings \cite{10.5555/3295222.3295349}. We train on the weakly-labeled source dataset for 200 epochs, followed by training on the union of weakly-labeled and external datasets for 40 CDL steps, each CDL step comprising 4 epochs. Additional information regarding hyper-parameters is provided in Supp. Material Section 8. \\
\textbf{Model Architecture}. Both ${P_m}$ and ${P_a}$ consist of a transformer encoder layer with four heads, followed by four fully connected layers, each consisting of 4096, 512, 32, and 1 neurons, respectively.  In both the models, for all the layers except the last, we use ReLU \cite{agarap2019deep} activation while for the last layer, we use Sigmoid activation. \\
\textbf{Evaluation Setup}. To reduce bias, we perform each experiment three times with different seeds and average the results. 
In open-set experiments, we repeat each experiment three times, using different sets of anomaly classes each time.\\
\textbf{Evaluation Metric}. Following previous works on UCF-Crime \cite{sultani2018real}, we adopt the frame-level area under the ROC curve (AUC) to evaluate on UCF-Crime. In line with previous works on XD-Violence \cite{Wu2020not}, we use the frame-level area under the Precision-Recall curve (PRAUC), also known as Average Precision (AP), to evaluate on XDV.

\begin{table}[t]
    \scriptsize
    \centering
    \begin{tabular}{@{}p{1.5cm}p{1.5cm}SSSS@{}}
        \toprule
        \textbf{} & \shortstack[c]{\textbf{Methods} \\ \text{}} & \shortstack{\textbf{Features} \\ \text{}} & \shortstack{\textbf{UCF} \\ \textbf{AUC(\%)}} & \shortstack{\textbf{UCF-R} \\ \textbf{AUC(\%)}} & \shortstack{\textbf{XDV} \\ \textbf{AP(\%)}} \\
        \midrule
         \multirow{3}{*}{\shortstack{Cross-Domain \\ (Unsup.)}} & \shortstack[l]{rGAN \text{\cite{10.1007/978-3-030-58558-7_8}}} & \shortstack{-} & \shortstack{$64.35^{\ast}$} & \shortstack{$65.19^{\ast}$} & \shortstack{$37.74^{\ast}$} \\
        & \shortstack[l]{MPN \text{\cite{Lv_2021_CVPR}}} & \shortstack{-} & \shortstack{$65.67^{\ast}$} & \shortstack{$67.98^{\ast}$} & \shortstack{$38.89^{\ast}$} \\
        & \shortstack[l]{zxVAD \text{\cite{Aich_2023_WACV}}} & \shortstack{-} & \shortstack{$68.74^{\dag}$} & \shortstack{$69.39^{\dag}$} & \shortstack{$40.68^{\dag}$} \\
        \midrule
         \multirow{5}{*}{\centering\shortstack{Non \\ Cross-\\ Domain}} & \shortstack[l]{Sultani \textit{et al.}\text{\cite{sultani2018real}}} & \shortstack{I3D} & \shortstack{80.70} & \shortstack{$84.63^{\ast}$} & \shortstack{$53.88^{\ast}$} \\
        & \shortstack[l]{MIST \text{\cite{feng2021mist}}} & \shortstack{I3D} & 82.30 & \text{$86.17^{\ast}$} & \text{$50.33^{\ast}$} \\
        & \shortstack[l]{RTFM \cite{tian2021weakly}} & \shortstack{I3D} & 84.03  & \text{$86.47^{\ast}$}  & \text{$37.30^{\ast}$}  \\ 
        & \shortstack[l]{S3R \cite{WuHCFL22}} & \shortstack{I3D} & 85.99  & \text{$87.11^{\ast}$}  & \text{$49.84^{\ast}$}    \\
        & \shortstack[l]{CU-Net \cite{Zhang_2023_CVPR}} & \shortstack{I3D} & 86.22 & \text{$88.15^{\ast}$} & \text{$37.98^{\ast}$} \\
        & \shortstack[l]{MGFN \cite{10.1609/aaai.v37i1.25112}} & \shortstack{I3D} & 86.98 &\text{$87.33^{\ast}$}  &   \text{$32.16^{\ast}$}    \\
        & \shortstack[l]{SSRL \cite{10.1007/978-3-031-19772-7_20}} &  \shortstack{I3D} & 87.43  & \text{$87.02^{\ast}$} & \text{$51.60^{\ast}$}   \\
        & \shortstack[l]{CLIP-TSA \cite{10222289}} & \shortstack{CLIP} & \shortstack{\textbf{87.58}} &  \shortstack{$73.20^{\ast}$}  & \shortstack{\text{$44.33^{\ast}$}}  \\
        & \shortstack[l]{Ours (No ext. data)} &\shortstack{CLIP} & 84.49 &  \shortstack{\text{89.96}}  & \shortstack{\text{58.13 }}  \\
        \midrule
        \multirow{2}{*}{\shortstack{Cross-Domain \\ (Weakly-Sup.)}} & \shortstack[l]{Ours (UCF + HACS)} & \shortstack{CLIP} & 84.63 &  \shortstack{\textbf{90.53 }}  & \shortstack{\text{65.14 }}  \\

        & \shortstack[l]{Ours (UCF + XDV)} & \shortstack{CLIP} & \shortstack{84.73} &  \shortstack{\text{90.26 }}  & \shortstack{\textbf{68.37 }}  \\
        \bottomrule
    \end{tabular}
    \caption{Comparison with prior works on XDV, considering UCF-Crime as the source data. Asterisk ($\ast$) indicates that evaluations were conducted by us using the official code. Dagger ($\dag$) indicates that evaluations were conducted by our implementation due to the lack of an official implementation.}
    \label{table:cdl-ucf}

\end{table}

\begin{table}[t]
    \scriptsize
    \centering
    \begin{tabularx}{\columnwidth}{l l@{} c c Xc@{}}
        \toprule
        \textbf{} & \textbf{Methods} & \textbf{Features} & \textbf{XDV AP(\%)} & \textbf{UCF-R AUC(\%)} \\
        \midrule
        \multirow{3}{*}{\shortstack{Cross-\\Domain \\(Unsup.)}} & rGAN \cite{10.1007/978-3-030-58558-7_8} & - & 40.10$^{\ast}$  & 59.82$^{\ast}$ \\
         & MPN \cite{Lv_2021_CVPR} & -  & 44.79$^{\ast}$ & 60.35$^{\ast}$ \\
         & zxVAD \cite{Aich_2023_WACV} & - & 47.53$^{\dag}$ & 63.61$^{\dag}$ \\
        \midrule
        \multirow{6}{*}{\shortstack{Non \\Cross-\\Domain}} & Sultani \textit{et al.}\cite{sultani2018real} & I3D & 73.20 & 71.23$^{\ast}$ \\
        & RTFM \cite{tian2021weakly} & I3D & 77.81 & 70.46$^{\ast}$ \\
         & MGFN \cite{10.1609/aaai.v37i1.25112} & I3D & 80.11 & 69.12$^{\ast}$ \\
         & S3R \cite{WuHCFL22} & I3D & 80.26 & 69.04$^{\ast}$ \\
         & CLIP-TSA \cite{10222289} & CLIP & \textbf{80.67} & 67.58$^{\ast}$ \\
         & Ours (No ext. data) & CLIP & 75.13 & 76.39 \\
        \midrule
        \multirow{2}{*}{\shortstack{Cross-Domain \\(Weakly-Sup.)}} & Ours (XDV + UCF) & CLIP & 77.04 & 88.06 \\
         & Ours (XDV + HACS) & CLIP & 78.61 & \textbf{88.50} \\
        \bottomrule    
    \end{tabularx}
    \caption{Comparison with prior works on UCF-Crime, considering XDV as the source data. Asterisk ($\ast$) indicates that evaluations were conducted by us using the official code. Dagger ($\dag$) indicates that evaluations were conducted by our implementation due to the lack of an official implementation.}
    \label{table:cdl-xdv}
\end{table}

\begin{table*}[t]
\small
    \centering
    \setlength{\tabcolsep}{4pt}
    \begin{tabular}{@{}c*{8}{S[table-format=2.2, table-number-alignment=center]}@{}}
        \toprule
         & \multicolumn{5}{c}{\textbf{UCF (AUC\%)}} & \multicolumn{2}{c}{\textbf{UCF-R (AUC\%)}} \\
         
        \boldmath{$c$} & \multicolumn{1}{c}{\textbf{Wu \textit{et al.}}\cite{Wu2020not}} & \multicolumn{1}{c}{\textbf{RTFM} \cite{tian2021weakly}} & \textbf{Zhu \textit{et al.} } \cite{10.1007/978-3-031-19830-4_23} & \textbf{Ours (w/o CDL)} & \textbf{Ours (CDL)} & \textbf{Ours (w/o CDL)} & \textbf{Ours (CDL)} \\
        
        \cmidrule(lr){1-1}  \cmidrule(lr){2-6}  \cmidrule(lr){7-8}        
        
        \textbf{1} & 73.22 & 75.91 & 76.73 & 75.17 & \textbf{77.45} & 84.32 & \textbf{85.39}\\ 
        \textbf{3} & 75.15 & 76.98 & 77.78 & 81.51 & \textbf{82.57} & 86.84 &  \textbf{87.69}\\
        \textbf{6} & 78.46 & 77.68 & 78.82 & 82.97 & \textbf{83.44} & 87.85 & \textbf{88.21}\\ 
        \textbf{9} & 79.96 & 79.55 & 80.14 & 83.02 & \textbf{83.37} & 89.22 &  \textbf{89.82}\\

        \bottomrule
    \end{tabular}%
    \caption{Comparison with other methods in Open-set setting on UCF-Crime dataset; $c$ denotes the no. of anomalous classes included for weakly-supervised training.\vspace{-10pt}}
    \label{table:open-set}
\end{table*}
\subsection{Noise in the Test Annotations of Benchmark Datasets} \label{subsec:noise}
Our manual inspection reveals that the frame-level testing annotations of the UCF-Crime (UCF) \cite{sultani2018real} and XD-Violence (XDV) \cite{Wu2020not} datasets, which are commonly used for benchmarking VAD models, exhibit significant noise. This noise largely stems from the fact that the original annotations do not consistently label the frames leading up to the primary anomalous events and their subsequent consequences as anomalous. 
For instance, in a video assigned a label like ``shooting'', we assert that frames showing the person holding the gun and frames illustrating the injured victim should also be marked as anomalous. This perspective aligns with the fundamental goal of VAD, which is to identify all anomalous frames within a video, irrespective of the video's primary label. However, it should also be noted that in the original annotations, for some videos, certain frames related to the video's primary anomaly label are also not marked anomalous.

To address this, we re-annotate the test set of UCF-Crime by assigning each video to three independent annotators. We then combine their annotations to generate more accurate frame-level labels. Compared to the original annotations where 7.58\% of the total frames are labeled as anomalous, the proposed annotations label 16.55\% of the total frames as anomalous. The proposed annotations are available here\footnote{\url{https://drive.google.com/drive/folders/1IVjQQFHXVcsaT63HUjpfk8C5KH6HsQ7t?usp=drive_link}}. We provide a comparison of the proposed and original annotations here\footnote{\url{https://rb.gy/4vkr1r}}. For the remainder of this paper, we refer the re-annotated test set of the UCF-Crime dataset as UCF-R. 

\subsection{Comparison with Prior Works} \label{subsec:comparison-prior}
\subsubsection{Cross-Domain Scenarios} \label{subsubsec:cross-domain}
While the UCF-Crime \cite{sultani2018real} and XD-Violence \cite{Wu2020not} datasets share similar definitions of what constitutes anomalies, that definition differs from those of smaller datasets like ShanghaiTech \cite{liu2018future}, CUHK-Avenue \cite{6751449}, UCSD Pedestrian \cite{4359353}, UBnormal  \cite{Acsintoae_CVPR_2022}, where anomalies are more subtle. For instance, running is considered anomalous in UBnormal but not in XD-Violence. Due to these divergent notions of anomalies across datasets, we conduct cross-domain experiments by simultaneously evaluating on the UCF-Crime and XD-Violence datasets, given their more aligned anomaly definitions. 

\textbf{UCF-Crime as the Weakly-Labeled Source Set, XDV as the Cross-Domain Set.} 
Table \ref{table:cdl-ucf} summarizes the results for this scenario.
First, we observe that the proposed method achieves state-of-the-art results on XDV and UCF-R even without utilizing any external data (without CDL). We believe this is due to the inductive bias of previous methods towards the noisy annotations of UCF-Crime. 
Next, we observe that the addition of external data, HACS and XDV, leads to a significant enhancement in the performance of the cross-domain dataset, XDV, by 11.26\% and 14.49\%, respectively, compared to the previous state-of-the-art baseline. Additionally, there is also a marginal improvement in the performance of the source set upon integration of external datasets.\\
\begin{figure*}[ht]
\small
  \centering
  \includegraphics[width=\textwidth]{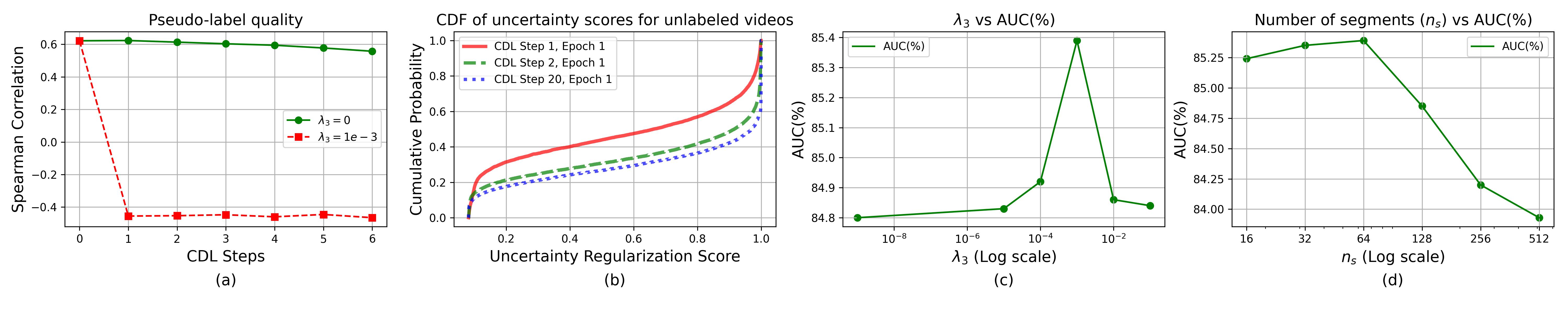}
   \caption{
   \textbf{(a)} Correlation between uncertainty scores and BCE loss computed between the estimated scores and ground truth. When $\lambda_3 = 1e-3$, as expected, a consistently high negative correlation emerges, demonstrating the effectiveness of the proposed uncertainty quantification method as a reliable proxy for pseudo-label quality. \textbf{(b)} Cumulative Distribution Function (CDF) plots illustrating the progression of average uncertainty regularization scores for each video during training. CDL step 20 has a higher concentration of scores around 1 compared to CDL step 2, while CDL step 2 has a higher concentration around 1 than CDL step 1. This suggests that, as training progresses, there is a higher tendency for scores to have elevated values, indicating more confident pseudo-label predictions.  
   \textbf{(c)} Ablation study on the coefficient of the cosine similarity loss term, $\lambda_3$.
   \textbf{(d)} Ablation study on the number of segments, $n_{s}$. 
  }
  \label{fig:figure-3}
\end{figure*}
\textbf{XDV as the Weakly-Labeled Source Set, UCF-Crime as the Cross-Domain Set.} Table \ref{table:cdl-xdv} summarizes the results for this scenario. Notably, the proposed method achieves state-of-the-art performance on the cross-domain dataset, UCF-R, even without the utilization of any external data during training. This is attributed to the simplicity of the proposed architecture compared to other baselines. The proposed architecture prevents overfitting to the source dataset, thereby increasing its generalizability to the cross-domain dataset.
Additionally, integrating external data further enhances performance on both the cross-domain and source sets. Specifically, leveraging the CDL framework with UCF-Crime and HACS as external datasets boosts UCF-R's AUC by 18.94\% and 19.39\% respectively, compared to previous state-of-the-art baselines. We also observe that the proposed method's performance is inferior on XDV. We attribute this to the noise in the annotations of XDV's test set.

These results highlight that the proposed CDL framework is capable of effectively exploiting external data with vast domain gaps to achieve a significant cross-domain generalization. It's noteworthy that the performance gain observed with the proposed CDL framework remains consistent across all tested datasets, suggesting that the performance improvement is not dependent on any specific source or external dataset.

\subsubsection{Open-Set Scenarios} \label{subsubsec:open-set}

In Table \ref{table:open-set}, we evaluate the proposed framework's performance on the UCF-Crime dataset in a realistic open-set scenario, where the model is evaluated on both, previously seen and unseen anomaly classes. To simulate this scenario, we randomly include $c$ anomalous classes in the weakly-labeled set, while the remaining anomalous classes are placed in the unlabeled set. In both the weakly-supervised source set and the unlabeled set, the number of normal videos equals the number of anomalous videos. We evaluate two model configurations; one trained solely on the weakly-labeled set (without CDL) and the other on the union of weakly-labeled and unlabeled sets using the CDL Framework.

On UCF-Crime, the proposed model, without CDL, surpasses the state-of-the-art baselines for $c>1$. This highlights its efficacy in open-set settings. While, with CDL, the model surpasses the baselines across all values of $c$ by a considerable margin.

For both UCF-Crime and UCF-R, when unlabeled data is incorporated, we observe a consistent performance gain across all values of $c$, suggesting the effectiveness of the CDL framework across varying amounts of weakly-labeled and unlabeled data. 

\subsection{Correlation between Uncertainty Scores and BCE Loss (Proxy to Label Quality)} \label{sec:correlation}
To assess the efficacy of the proposed uncertainty quantification method as a proxy for pseudo-label quality, we compute the non-parametric Spearman correlation between estimated uncertainty regularization scores and BCE loss between the predicted pseudo-labels and the corresponding ground truths. For this experiment, we consider UCF-Crime as the weakly-labeled source set and XDV as the external set. In Figure \ref{fig:figure-3}(a), with $\lambda_{3} = 1e-3$, CDL step 1 onwards, a consistently high negative correlation (-0.46 in CDL step 6, with a p-value $\textless$ 1e-5) emerges, indicating the robustness of the proposed uncertainty quantification framework. Conversely, setting $\lambda_{3}$ to 0 results in a sustained positive correlation, signifying sub-optimal pseudo-labels in the absence of cosine similarity loss term.
\subsection{Progression of Uncertainty Scores} \label{subsec:progression}
To assess the evolution of uncertainty regularization scores through the training process, in Figure \ref{fig:figure-3}(b), we plot the Cumulative Distribution Function (CDF) of average uncertainty regularization scores for external videos across the first epoch of three different CDL steps. We conduct this experiment considering UCF-Crime as the weakly-labeled source set and XDV as the external set. We observe that in CDL step 1, 16.65\% of the uncertainty scores fall within the range [0, 0.1]. As training progresses to CDL steps 2 and 20, this proportion decreases to 13.06\% and 11.39\%, respectively. Meanwhile, the proportion of uncertainty scores in the range [0.9, 1] increases from 35.11\% in CDL step 1 to 56.70\% in CDL step 2 and further to 57.68\% in CDL step 20.  This trend indicates a discernible shift towards higher uncertainty scores as training progresses, suggesting an improvement in model confidence due to increased pseudo-label quality.

\subsection{Ablation Studies and Hyper-parameter Analysis} \label{sec:ablation-studies}

\begin{figure}[t]
  \centering
  \includegraphics[width=0.4\textwidth]{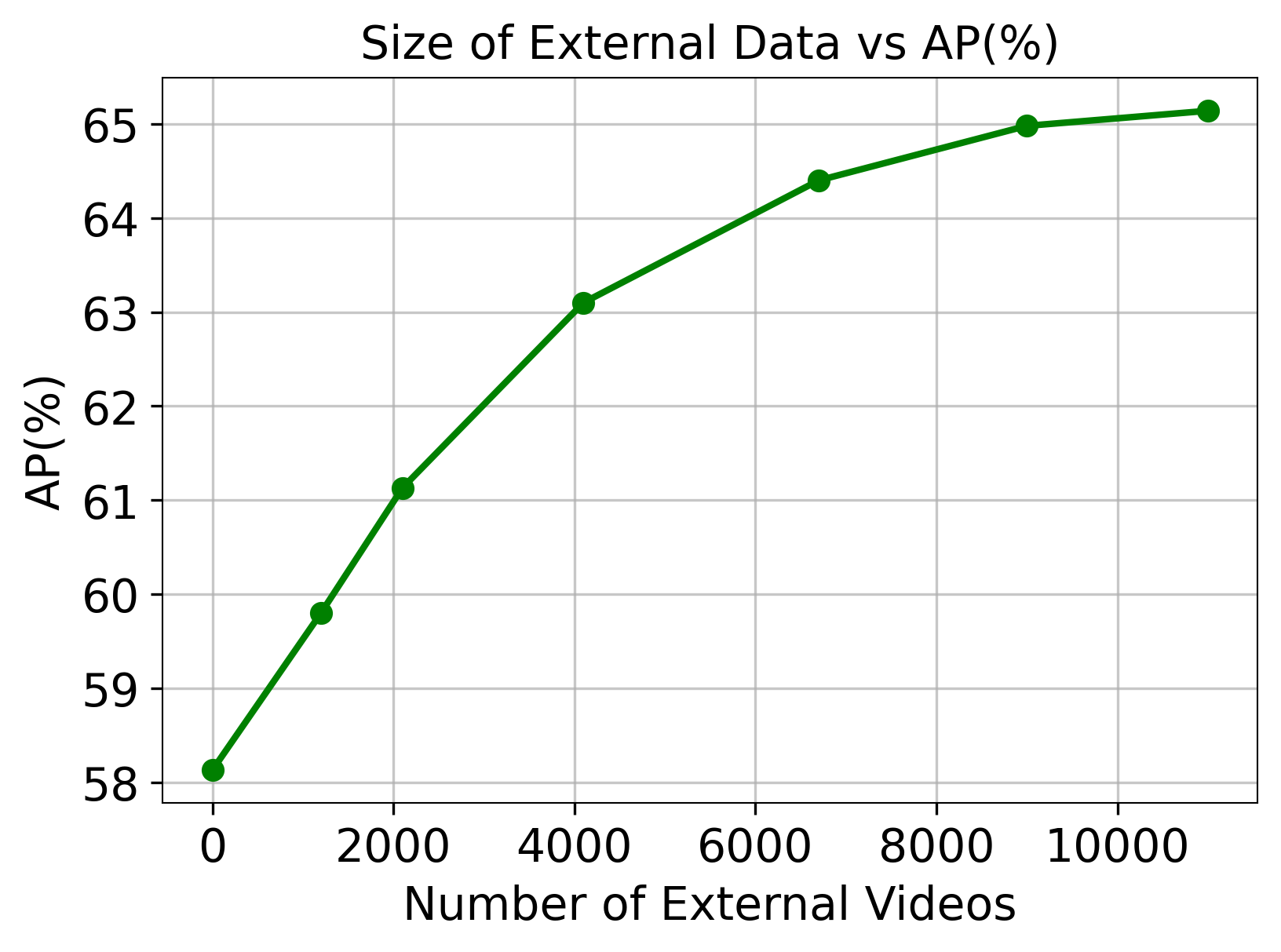} 
  \caption{Ablation study on the impact of the size of external data.}
  \label{fig:data-vs-ap}
\end{figure}

For the sake of consistency, we conduct all ablation studies on UCF-Crime in an open-set setting, with $c = 1$. 
However, it should be noted that for different training setups, hyper-parameters are tuned separately as well.\\
\textbf{Impact of Various Components of the CDL Framework}. 
We assess the effectiveness of each component of the CDL framework by adding them sequentially. The results are summarized in Table \ref{table:ablation}. 
We consider training on $c = 1$ anomaly class in a weakly-supervised fashion as our baseline. The remaining $c-1$ anomalous classes are placed in the external set.         
We first observe that integrating external data into the source set without accounting for pseudo-label uncertainty ($S^{i, j} = 1, \forall i, j$) and without minimizing cosine similarity between representations ($\lambda_3 = 0$) yields a 0.35\% gain in AUC, highlighting the effectiveness of external data in improving the model's performance. Next, we study the impact of uncertainty-aware integration of external data, \ie, adaptively reweighing the prediction bias of external data with the computed uncertainty values and with $\lambda_3$ set to 0. This results in a gain of 0.13\% in AUC, demonstrating the superiority of uncertainty-driven integration compared to the standard integration. Finally, we assess the impact of adding the cosine similarity loss term during uncertainty-aware training. This further leads to a significant boost of 0.59\%, validating its effectiveness. \\
\textbf{Impact of Cosine Similarity Loss}. In Figure \ref{fig:figure-3}(c), we explore the impact of varying the coefficient of the cosine similarity loss on the model's performance. We observe a gradual increase in AUC as $\lambda_3$ increases from 1e-9 to 1e-3. This could be due to the effect of cosine similarity loss getting more pronounced with higher values of $\lambda_3$. However, beyond 1e-3, there is a rapid decline in AUC, likely due to the dominance of the cosine similarity loss over other losses when its coefficient is high. Therefore, we select 1e-3 as the optimal choice for $\lambda_3$.\\
\begin{table}[t]
    \centering
    \footnotesize
    \setlength{\tabcolsep}{2pt} 
    \begin{tabular}{>{\centering\arraybackslash}p{2cm} *{4}{>{\centering\arraybackslash}c}}
        \toprule
        \textbf{External data} & \textbf{Uncertainty Coeff.} & \textbf{Cos. Similarity Loss} & \textbf{AUC(\%)} \\
        \midrule
         \ding{55} & \ding{55} & \ding{55} & 84.32 \\
         \ding{51} & \ding{55} & \ding{55} & 84.67 \\
         \ding{51} & \ding{51} & \ding{55} & 84.80 \\
         \ding{51} & \ding{51} & \ding{51} & 85.39 \\
        \bottomrule
    \end{tabular}
    \caption{Ablation study of various components on the UCF-R dataset in an open-set setting ($c=1$).}
    \label{table:ablation}
\end{table}
\textbf{Impact of Number of Segments}. In Figure \ref{fig:figure-3}(d), we observe that the performance consistently improves as no. of segments, $n_s$, increases from 16 to 64, but it begins to decline rapidly afterward. Therefore, we set $n_s$ as 64.\\
\textbf{Impact of the Size of External Data}. To determine the optimal number of unlabeled external videos from the HACS dataset to integrate into the weakly-labeled training set of UCF-Crime, we conduct an ablation study, depicted in Figure \ref{fig:data-vs-ap}. We observe that increasing the size of the external set increases the performance on XDV. However, this increase tends to plateau after the inclusion of 11,000 videos. Consequently, we do not include additional videos beyond the 11,000 threshold.

\section{Conclusion}
In this work, we demonstrated the effectiveness of integrating external, unlabeled data with weakly-labeled source data to enhance the cross-domain generalization of VAD models. To enable this integration, we proposed a weakly-supervised CDL (Cross-Domain Learning) framework that adaptively minimizes the prediction bias on external data by scaling it with the prediction variance, which serves as an uncertainty regularization score. The proposed method outperforms baseline models significantly in cross-domain and open-set settings while retaining competitive performance in in-domain settings. 

\section*{Acknowledgement}
This work was supported in part by U.S. NIH grants R01GM134020 and P41GM103712, NSF grants DBI-1949629, DBI-2238093, IIS-2007595, IIS-2211597, and MCB-2205148. This work was supported in part by Oracle Cloud credits and related resources provided by Oracle for Research, and the computational resources support from AMD HPC Fund. We thank Eshaan Mandal and Bhavay Malhotra for their assistance, which has been instrumental in completing this work.

{
    \small
    \bibliographystyle{ieeenat_fullname}
    \bibliography{arxiv}
}

\clearpage
\maketitlesupplementary

\section{Revisiting Multiple Instance Learning}
\label{supp-sec:mil}
Since acquiring frame-level labels requires significant time and effort, following Sultani \textit{et al.} \cite{sultani2018real}, we use Multiple Instance Learning (MIL) to train the classifiers using weakly-supervised video-level labels. By dividing a video (bag) into multiple temporal non-overlapping segments (instances) and encouraging anomalous video segments to have higher anomaly scores as compared to the normal segments, they formulate anomaly detection as a regression problem.

The multiple instance ranking objective function is given by:

\begin{equation}
\max_{\substack{X^i \in \mathcal{D}_l^a \\ 1 \leq j \leq n_s}} F(X^{i, j}|\theta) > \max_{\substack{X^i \in \mathcal{D}_l^n \\ 1 \leq j \leq n_s}} F(X^{i, j}|\theta),
\end{equation}
where $\mathcal{D}_l^a = \{
(X, Y) \in \mathcal{D}_l: Y=1\}$ and $\mathcal{D}_l^n = \{
(X, Y) \in \mathcal{D}_l: Y=0\}$ are the set of abnormal and normal videos, respectively and max is taken over all video segments in a bag.

Instead of ranking every segment of the positive and negative bags, ranking is enforced on one segment from each bag, having the highest anomaly score.
The overall loss function, $\mathcal{L}_{\text{rank}}$, for a pair of abnormal and normal videos, is given by:
\begin{equation}
\begin{split}
\mathcal{L}_{\text{rank}} = &\max(0, 1 - \max_{\substack{X^i \in \mathcal{D}_l^a \\ 1 \leq j \leq n_s}} F(X^{i, j}|\theta) + \max_{\substack{X^i \in \mathcal{D}_l^n \\ 1 \leq j \leq n_s}} F(X^{i, j}|\theta)) \\
&+ \lambda_{1} \mathcal{L}_{\text{Ts}} + \lambda_{2} \mathcal{L}_{\text{Sp}},
\end{split}
\end{equation}
where $\mathcal{L}_{\text{Ts}}$
is the temporal smoothness constraint,
and $\mathcal{L}_{\text{Sp}}$ is the sparsity constraint.

\section{Datasets} \label{sup:datasets}
\textbf{UCF-Crime} \cite{sultani2018real}: This is a large-scale VAD dataset having a total duration of 128 hours. It contains long and untrimmed real-world surveillance videos across 13 realistic anomaly categories that are specifically chosen due to their significant impact on public safety. The dataset comprises 1610 weakly-labeled training videos and 290 test videos annotated at the frame level. \\
\textbf{XD-Violence (XDV)} \cite{Wu2020not}: This is a large-scale and multi-scene audio-visual dataset for violence detection, having a total duration of 217 hours. Its long and untrimmed videos are collected from movies, games, and in-the-wild scenarios, with anomalies spread over 6 categories. It comprises 3954 weakly-labeled training videos and 800 test videos annotated at the frame level.\\
\textbf{HACS} \cite{zhao2019hacs}: This is a large-scale dataset for human action recognition, sourced from YouTube. It features 200 action classes across 140K segments on 50K videos. Due to its diverse range of actions, larger size, and longer video durations compared to other video datasets such as UCF-101, Kinetics, and ActivityNet, we use a subset of 11K videos from HACS Segments as external, unlabeled data.

\section{Implementation Details}   \label{sup:details}
To ensure consistency and gradient stability, while training on ${\mathcal{D}_l} \cup {\mathcal{D}_u}$, each mini-batch consists of an equal number of samples from ${\mathcal{D}_l}$ and ${\mathcal{D}_u}$. Since the computation of $\mathcal{L}_{\text{rank}}$ necessitates pairs of abnormal and normal videos, each labeled sample within the mini-batch comprises a pair of anomalous and normal videos. All the experiments were conducted on an NVIDIA RTX A5000 24 GB GPU. For the experiments using UCF-Crime as the weakly-labeled data, we set the batch size to 64, and for the experiments using XD-Violence as the weakly-labeled data, we set the batch size to 32.
In all our experiments except the open-set, we set $n_s$ to 64, $\tau$ to 1.25, $\lambda_1$ to 5e-3, $\lambda_2$ to 1e-3, $\lambda_3$ to 1e-3. 
We set $\lambda_4$ to 2000 for UCF+HACS and UCF+XDV, 1250 for XDV+HACS, and 700 for XDV+UCF.
For all our experiments, we use the Adam optimizer with a weight decay of 1e-3. 
For the fully connected layers, we use a learning rate of 5e-4 when UCF-Crime is used as the weakly-labeled dataset and a learning rate of 1e-4 when XDV is used as the weakly-labeled dataset.
For the transformer encoder layers, we use a learning rate of 3e-5 when UCF-Crime is used as the weakly-labeled dataset and a learning rate of 5e-5 when XDV is used as the weakly-labeled dataset. 
In all our experiments, we explicitly encode positional information in the segments using sinusoidal positional encodings \cite{10.5555/3295222.3295349}. We train on the weakly-labeled source dataset for 200 epochs, followed by training on the union of weakly-labeled and external datasets for 40 CDL steps, each CDL step comprising 4 epochs.
Due to the finer granularity and semantic richness inherent in CLIP features, we choose to use CLIP features during inference.

\begin{figure*}[!t]
  \centering
  \includegraphics[width=0.65\textwidth]{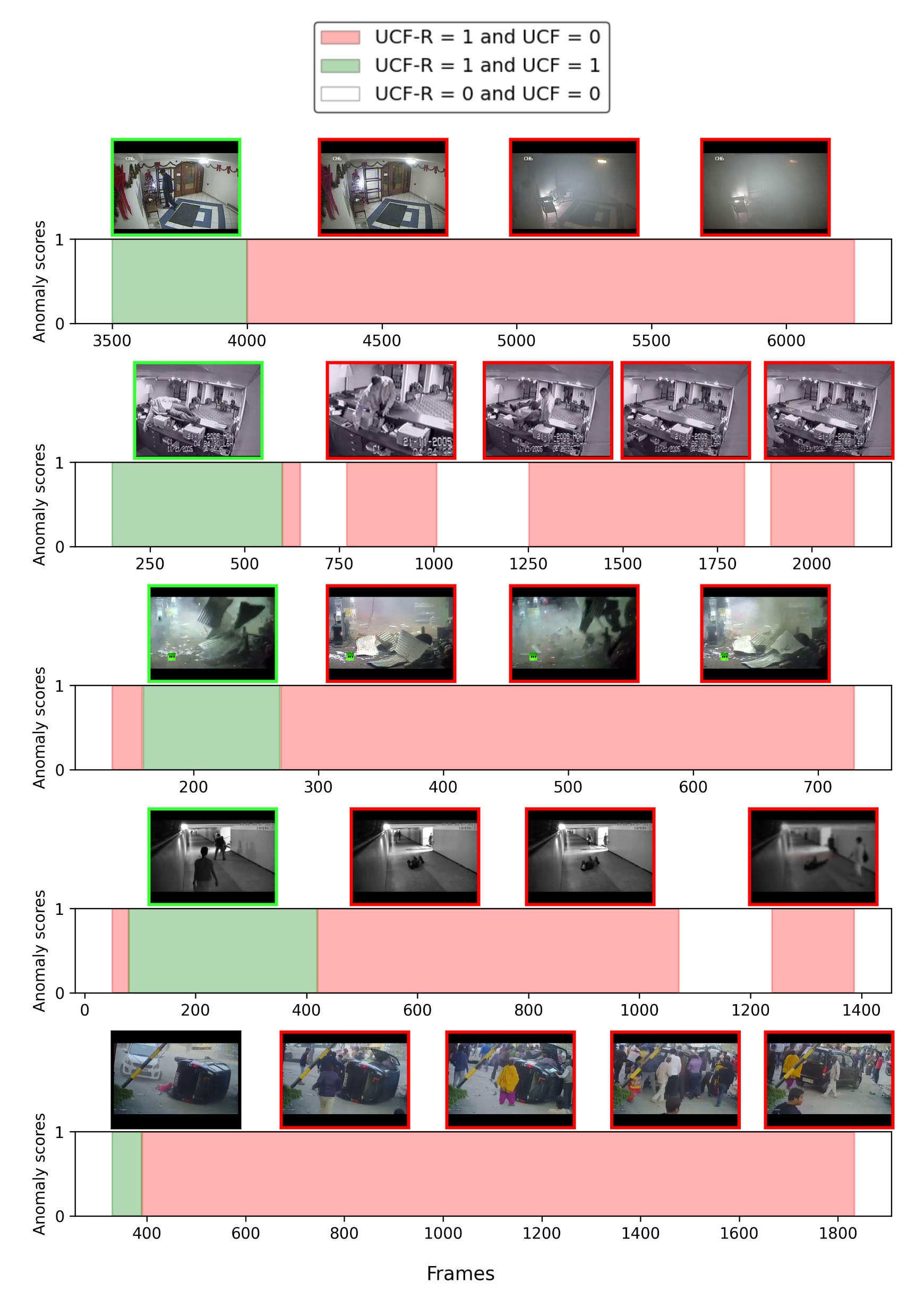}
  \caption{A comparison between the original annotations (UCF) and the proposed annotations (UCF-R). The green region represents frames labeled as anomalous by both the original and proposed annotations. The red region indicates frames labeled as anomalous by the proposed annotations but not by the original annotations. The unshaded (white) region denotes normal frames. For instance, in the first row, while the original annotations just label frames depicting arson (a person setting the Christmas tree on fire) as anomalous, UCF-R also labels the frames depicting the fire and smoke following arson as anomalous.} 
  \label{fig:comparison}
\end{figure*}

\section{Comparison with Unsupervised Baselines in Open-Set Settings}
Table \ref{tab:extended-open-set} depicts that the proposed method outperforms all the baselines in open-set settings on the UCF-Crime dataset by a large margin. As expected, all the weakly-supervised methods outperform the unsupervised methods, even when a small subset of the data is used for weakly-supervised training. This highlights the necessity of incorporating weak labels during training. Since a direct comparison of the proposed weakly-supervised framework with unsupervised methods is not fair, we did not include unsupervised baselines in Table 4.

\begin{table*}[ht]
    \centering
    \caption{Comparison with prior works in open-set setting on UCF-Crime dataset; $c$ denotes the number of anomalous classes included for weakly-supervised training. The values represent AUC (\%).}
    \begin{small}
    \setlength{\tabcolsep}{5pt}
            \begin{tabular}{p{1.7cm} p{3cm} *{5}{c}}
                \toprule
                & \textbf{$c$}    & \textbf{0} & \textbf{1} & \textbf{3} & \textbf{6} & \textbf{9} \\
                \midrule
                \multirow{5}{*}{Unsup.} & Conv-AE~\cite{hasan2015context} & 
                50.60 & - & - & - &- \\
                & Sohrab \etal~\cite{sohrab2018subspace} &
                58.50 & - & - & - & - \\
                & Lu \etal~\cite{6751449} & 
                65.51 & - & - & - &- \\
                & BODS~\cite{9008531} & 
                68.26 & - & - & - &- \\
                & GODS~\cite{9008531} & 
                70.46 & - & - & - &- \\
                \midrule
                \multirow{6}{*}{Weakly-Sup.} & Wu \etal\cite{Wu2020not} (offline) & -& 73.22 & 75.15 &  78.46   &  79.96 \\
                & Wu \etal\cite{Wu2020not} (online) & -& 73.78 & 74.64 &  77.84 &  79.11 \\
                & RTFM \cite{tian2021weakly} & - & 75.91 & 76.98 &77.68 & 79.55 \\        
                & Zhu \etal \cite{10.1007/978-3-031-19830-4_23} &  - & 76.73 & 77.78 & 78.82 & 80.14 \\
                & Ours (w/o CDL) &  - & 75.17 & 81.51 & 82.97 & 83.02 \\
                & Ours &  - & 77.45 & 82.57 & 83.44 & 83.37 \\
                \bottomrule
            \end{tabular}
            \label{tab:extended-open-set}
    \end{small}
\end{table*}

 \section{Comparison of the Original and Proposed Annotations for UCF-Crime Dataset}
Figure \ref{fig:comparison} illustrates a subset of instances from the UCF-Crime's test set where the original annotations do not label frames as anomalous, despite their actual anomalous nature. 
We also provide a comparison of the proposed and original annotations superimposed on the videos at this link: \url{https://rb.gy/4vkr1r}.

\section{Limitations}
Similar to some recent weakly-supervised VAD works \cite{feng2021mist, Li_Liu_Jiao_2022, Zhang_2023_CVPR}, the training process of the proposed CDL framework involves two stages. Consequently, the training does not operate in an end-to-end manner. This incurs additional complexity and challenges for training the model in real-world applications. However, since the generalization obtained using this multi-stage training is significant, the complex training setup of the multi-stage framework is reasonable. Nonetheless, developing end-to-end training frameworks would be an important direction for future research. This can facilitate the advancement of anomaly detection approaches for real-world applications, particularly the ones with limited training budgets.

Additionally, the cross-domain performance in case of drastic distribution shifts between the source and target domains may be hindered. For instance, a model primarily trained on videos from stationary surveillance cameras may not effectively work on videos with rapidly evolving scenes from car dashcams.
This is mainly because the uncertainty-based reweighing approach in our framework aims to select samples from the external set that are similar to the source domain. In case of drastic shifts between the two domains, finding informative samples from the target domain would not be trivial.

\end{document}